\documentclass[runningheads]{llncs}
\usepackage{enumitem}

\usepackage[utf8]{inputenc} % allow utf-8 input
\usepackage[T1]{fontenc}    % use 8-bit T1 fonts
\usepackage{hyperref}       % hyperlinks
\usepackage{url}            % simple URL typesetting
\usepackage{booktabs}       % professional-quality tables
\usepackage{amsfonts}       % blackboard math symbols
\usepackage{nicefrac}       % compact symbols for 1/2, etc.
\usepackage{microtype}      % microtypography
\usepackage{lipsum}
\usepackage{array} % for \arraybackslash
\usepackage{fancyhdr}       % header
\usepackage{graphicx}       % graphics
\graphicspath{{media/}}     % organize your images and other figures under media/ folder
\usepackage{booktabs}
\usepackage{tabularx}
\usepackage{makecell} % for manual line breaks in headers
\newcolumntype{L}{>{\raggedright\arraybackslash}X}
\usepackage{amsmath}
\usepackage{tcolorbox}
\tcbuselibrary{breakable}
\newtcolorbox{tipsbox}[1]{% 1 = number of arguments
    breakable,
    title={\textbf{Tips:} #1},          % use the argument as the title
    colback=blue!5,
    colframe=blue!60,
    colbacktitle=blue!20,
    coltitle=black,
    boxrule=1pt,
    fonttitle=\bfseries,
    arc=2mm,
    left=6pt, right=6pt, top=6pt, bottom=6pt
}

\newtcolorbox{defbox}[1]{% 1 = number of arguments
    breakable,
    title={\textbf{Definition:} #1},          % use the argument as the title
    colback=orange!5,
    colframe=orange!60,
    colbacktitle=orange!20,
    coltitle=black,
    boxrule=1pt,
    fonttitle=\bfseries,
    arc=2mm,
    left=6pt, right=6pt, top=6pt, bottom=6pt
}

\usepackage{orcidlink} % in the preamble
  
\title{Best Practices for Machine Learning Experimentation in Scientific Applications
}

\date{\today}

\author{Umberto Michelucci\inst{1}\orcidlink{0000-0002-6060-5365} \and
Francesca Venturini\inst{2}\orcidlink{0000-0003-2562-9932}
}
\authorrunning{U. Michelucci et al.}
\institute{Computer Science Department, Lucerne University of Applied Sciences and Arts,\\
Luzern 6002, Switzerland\\
\email{umberto.michelucci@hslu.ch}
\and
Institute of Applied Mathematics and Physics,\\
ZHAW - Zurich University of Applied Sciences,\\
Winterthur 8400, Zurich, Switzerland\\
\email{vent@zhaw.ch} 
}

\begin{document}

\maketitle

\begin{abstract}
Machine learning (ML) is increasingly adopted in scientific research, yet the quality and reliability of results often depend on how experiments are designed and documented. Poor baselines, inconsistent preprocessing, or insufficient validation can lead to misleading conclusions about model performance. This paper presents a practical and structured guide for conducting ML experiments in scientific applications, focussing on reproducibility, fair comparison, and transparent reporting. We outline a step-by-step workflow, from dataset preparation to model selection and evaluation, and propose metrics that account for overfitting and instability across validation folds, including the Logarithmic Overfitting Ratio (LOR) and the Composite Overfitting Score (COS). Through recommended practices and example reporting formats, this work aims to support researchers in establishing robust baselines and drawing valid evidence-based insights from ML models applied to scientific problems.

\end{abstract}

%\tableofcontents

% keywords can be removed
%\keywords{Machine Learning \and Baseline \and Research \and Experiments}

\section{Introduction}

When starting a new machine learning (ML) project, one of the most critical steps is to create a solid and well-documented \textbf{baseline}. A baseline provides a reference point against which more complex models such as deep neural networks can be objectively compared. Establishing it in the right way ensures that any subsequent improvement truly reflects the model design and not differences in data processing or evaluation.

This document presents a step-by-step procedure for building a sensible baseline, from dataset preparation to model evaluation and reporting. Each step emphasises reproducibility, transparency, and fair comparison between approaches. Table~\ref{tab:summary_baseline} at the end of this section summarises the main recommendations and good practices. In Table \ref{tab:ml_algorithms} you find an overview of the most commonly used non-deep learning algorithms to try. åAn example of a possible \textit{output} table can be found in Table \ref{tab:resultsexample}.
\begin{table}[h!]
\centering
\renewcommand{\arraystretch}{1.3}
\setlength{\tabcolsep}{8pt}
\begin{tabular}{l p{7cm}}
\toprule
\textbf{Step} & \textbf{Main Goals and Practical Tips} \\
\midrule
\textbf{Data Preparation} & Clean, normalize, and save each dataset version. Document all transformations and check for outliers visually. \\
\textbf{Classical Baseline} & Test simple models (Linear/Logistic Regression, Trees, Random Forest, etc.). Keep hyperparameters simple and record baseline metrics. \\
\textbf{Cross-Validation} & Use $k$-Fold or Monte Carlo CV. Record mean and std of metrics, and check for overfitting patterns. \\
\textbf{Deep Learning Models} & Introduce simple neural nets. Use regularization and early stopping. Ensure same preprocessing as classical models. \\
\textbf{Model Selection} & Exclude non-learning or overfitting models. Keep reproducible, stable, interpretable ones. \\
\textbf{Reporting Results} & Summarize in a table (metrics ± SD, both for training and test). Include preprocessing details and highlight best performing models. \\
\bottomrule
\end{tabular}
\vspace{0.2cm}
\caption{Summary of the main steps and recommendations for building a solid ML baseline.}
\label{tab:summary_baseline}
\end{table}

The  steps defined in the following sections should be followed when designing machine learning experiments. 

\subsection{Design of Experiments}

An ML experiment is a controlled study in which you train and evaluate one or more ML modeltypes under specified conditions to answer a concrete question. 
\begin{defbox}{ML Experiment}
    An ML experiment is a controlled study in which you train and evaluate one or more ML model types under specified conditions to answer a concrete question. 
\end{defbox}
For example, imagine that you are trying to classify some input (for example Raman spectra) to predict an output (for example, a concentration of some chemical components), then an experiment is made of the following main components:
    \begin{enumerate}
        \item dataset with preprocessing steps (more on that later);
        \item machine learning model type (for example, logistic regression or linear regression);
        \item machine learning model instance (for example, support vector classifier with a regularisation parameter $C=10$), in other words, the model with a chosen set of parameters ($\alpha$ in ridge regression, regularisation parameter; 
        $C$ in SVC, etc.);
        \item metrics on the training and validation datasets;
        \item final summary of results (summarised in a table).

\begin{defbox}{Model Type vs. Model Instance}
A \textbf{model type} refers to the generic algorithmic family used for learning from data 
(e.g.\ Logistic Regression, Support Vector Machine, Random Forest). 
It defines the overall structure and assumptions of the learning process.

A \textbf{model instance} is a concrete realisation of a model type, obtained by 
choosing a specific set of hyperparameters (e.g.\ Logistic Regression with 
$\ell_2$ regularisation strength $C = 10$, or SVM with RBF kernel and $\gamma = 0.1$). 
It is the actual model that is trained and evaluated in an experiment.
\end{defbox}

    \noindent \textbf{IMPORTANT}: An experiment is defined by \textbf{one single set} of all parameters/preprocessing steps.
    \end{enumerate}
    To design one or multiple experiments, it is useful to fill out the following matrix (which will be useful later on for documentation). In Table \ref{tab:experiments} you see some examples.
\begin{table}[h!]
    \centering
    \footnotesize                        % a bit smaller
    \setlength{\tabcolsep}{2pt}          % less horizontal padding
    \renewcommand{\arraystretch}{1.2}    % nicer row height

    \begin{tabularx}{\textwidth}{
        l      % Exp. ID
        l      % Task
        L      % Preprocessing
        L      % Normalisation
        L      % Model Instance
        L      % Metrics
        l      % Dataset
        L      % Notes
    }
        \toprule
        \textbf{Exp. ID} &
        \textbf{Task} &
        \textbf{Preproc.} &
        \textbf{Normal.} &
        \textbf{Instance} &
        \textbf{Metrics} &
        \textbf{Dataset} &
        \textbf{Notes} \\
        \midrule

        EX1 & Classification &
        Raw &
        $\text{max} = 1$ &
        Decision Tree &
        Accuracy, F1 &
        v1 &
        First quick baseline \\

        EX2 & Classification &
        Baseline removed &
        None &
        Random Forest &
        Accuracy, F1 &
        v2 &
        More complex model \\
        \bottomrule
    \end{tabularx}
    \\[0.2cm]

    % \caption{Example of the table that you can fill to plan the experiments you want to do.}
    % \label{tab:experiments}
    \caption{Example of the table that you can fill to plan the experiments you want to do.}
    \label{tab:experiments}
\end{table}   
%\end{table}

    Once you have performed all your experiments, your results should be put in a similar table. You can see how this table might look like in Table \ref{tab:resultsexample} with the MAE as a metric. If you have more metrics you will need more columns.

\begin{table}[h!]
    \centering
    \footnotesize                        % a bit smaller
    \setlength{\tabcolsep}{2pt}          % less horizontal padding
    \renewcommand{\arraystretch}{1.2}    % nicer row height

    \begin{tabularx}{\textwidth}{
        l      % Exp. ID
        l      % Task
        L      % Preprocessing
        L      % Normalisation
        L      % Model Instance
        L      % Metrics
        l      % Dataset
        L      % Notes
    }
        \toprule
        \textbf{Exp. ID} &
        \textbf{Model} &
        \textbf{Preproc.} &
        \textbf{Normal.} &
        \textbf{MAE$\pm \sigma$ (train)}&
        \textbf{MAE$\pm \sigma$ (test)} &
        \textbf{LOR} &
        \textbf{COS} \\
        \midrule

        EX1 & Decision Tree &
        Raw &
        $\text{max} = 1$ &
        $3.4\pm 0.2$ &
        $3.0\pm 0.3$ &
        0.054 &
        0.9 \\

        EX2  & Random Forest &
        Basedline Removed &
        $\text{max} = 1$ &
        $3.5\pm 0.1$ &
        $2.6\pm 0.4$ &
        0.13 &
        0.80 \\
        \bottomrule
    \end{tabularx}
\\[0.2cm]
    % \caption{Example of the table that you can fill to plan the experiments you want to do.}
    % \label{tab:experiments}
    \caption{Example of the table to show the different experiments. The metric reported is the Mean Absolute Error (MAE) but could be something else for you, for example the MSE, or the accuracy, or something completely different.}
    \label{tab:resultsexample}
\end{table}

    \subsection{Dataset Preparation} While designing experiments you have to decide how to prepare your data. Put the information in the two columns in Table \ref{tab:experiments} in the columns \textit{Preprocessing} and \textit{Normalisation}.

    \begin{tipsbox}{Dataset preparation}
    \begin{enumerate}[label=(\roman*)]
        \item Always keep the raw data intact; create derived, preprocessed versions.
        \item Document every transformation (normalisation, outlier removal, encoding, etc.).
        \item Save datasets at each stage as separate pickle files:
        data\_raw.pkl, data\_normalized\_v1.pkl, data\_cleaned\_v1.pkl, etc.
        % \item Visualize your distributions before and after normalization or cleaning to ensure no distortions.
    \end{enumerate}
    \end{tipsbox}

    \subsection{Choosing classical models for the experiments} In this phase you decide which model types and their instances you want to use, train them and record the results. Put this information in Table \ref{tab:experiments} in the columns \textit{Model Instance} for which model you want to test and in the column \textit{metrics} for which metric you have chosen.
    
        \begin{tipsbox}{Choosing the right model type}
        \begin{enumerate}[label=(\roman*)]
        \item Start simple: linear or logistic regression often give surprising insight into data  (see Table \ref{tab:ml_algorithms} for a list of the most commonly used algorithms).
        \item Use simple models to check data leakage and confirm label consistency.
        \item Keep hyperparameters minimal at this stage; focus on understanding baseline performance.
        \item Record training time and model simplicity (helps later justify complexity of advanced models).
        \item \textbf{Always record training and test metrics} to check for overfitting.
    \end{enumerate}
    \end{tipsbox}

\begin{table}[h!]
    \centering
    \begin{tabular}{>{\centering\arraybackslash}p{3cm} p{5cm} p{4cm}}
    \toprule
    \textbf{Algorithm Name} & \textbf{Short Description} & \textbf{Python (Scikit-learn)} \\
    \midrule
    \multicolumn{3}{c}{\textbf{Regression Algorithms}} \\
    \midrule
    \textbf{Linear Regression} & Models the relationship between a scalar dependent variable and explanatory variables using a linear equation. & \texttt{from sklearn.linear\_model import LinearRegression} \\
    \midrule
    \textbf{Random Forest Regressor} & An ensemble method that constructs multiple decision trees and outputs the average prediction of the individual trees. & \texttt{from sklearn.ensemble import RandomForestRegressor} \\
    \midrule
    \textbf{AdaBoost Regressor} & An ensemble boosting method that combines multiple weak regressors (e.g., decision trees) to create a strong predictor. & \texttt{from sklearn.ensemble import AdaBoostRegressor} \\
    \midrule
    \textbf{Support Vector Machine (SVR)} & Uses kernel functions to find a hyperplane that fits the data while keeping errors within a maximum margin ($\epsilon$). & \texttt{from sklearn.svm import SVR} \\
    \midrule
    \multicolumn{3}{c}{\textbf{Classification Algorithms}} \\
    \midrule
    \textbf{Logistic Regression} & Uses the logistic function to model a binary dependent variable; estimates the probability of an event occurring. & \texttt{from sklearn.linear\_model import LogisticRegression} \\
    \midrule
    \textbf{Support Vector Machine (SVC)} & Finds an optimal hyperplane in an $N$-dimensional space that distinctly classifies data points and maximizes the margin. & \texttt{from sklearn.svm import SVC} \\
    \midrule
    \textbf{K-Nearest Neighbors (KNN)} & A non-parametric algorithm that classifies a new data point based on the majority class of its $k$ nearest neighbors. & \texttt{from sklearn.neighbors import KNeighborsClassifier} \\
    \midrule
    \textbf{Random Forest} & An ensemble method that constructs multiple decision trees and outputs the average prediction of the individual trees. & \texttt{from sklearn.ensemble import RandomForestClassifier} \\
    \bottomrule
    \end{tabular}
    \\[0.2cm]
    \caption{Often Used Machine Learning Algorithms for Regression and Classification}
    \label{tab:ml_algorithms}
\end{table}

    \subsection{Cross-Validation (CV) and Evaluation} Always use a proper validation model. You should use: Leave One Out if you have very small datasets (say 20-30 elements). $k$-Fold or Monte-Carlo CV if you have slightly larger datasets. The standard deviation in the metric columns in Table \ref{tab:resultsexample} is the standard deviation over the different folds (in $k$-fold CV) or over the multiple splits (in Monte Carlo CV).

    \begin{tipsbox}{Cross validation}
    \begin{enumerate}[label=(\roman*)]
        \item Prefer Monte Carlo CV; $k$-Fold CV is good for larger, stable datasets.
        \item Compute and log both mean and standard deviation of your metrics.
        \item Track signs of overfitting: large gaps between train/test performance indicate data leakage or model variance. Always calcualte and report LOR (see below) and COS (see below).
        \item Store evaluation results in structured form (CSV, DataFrame, etc.) for reproducibility.
        \item \textbf{Always} plot loss function vs. epochs to check learning (for deep learning). 
        \item \textbf{For Classification}: always use confusion matrix and not only accuracy.
        \item \textbf{For Regression}: always plot prediction vs. true values and calculate $R^2$ values not only MAE.
    \end{enumerate}
    \end{tipsbox}
    CV is done to check generalisation properties of models, meaning how they behave on new data. In more simple terms, it is done to check if models overfit (for a more nuanced discussion about overfitting you can check Chapter 7 in \cite{michelucci2024fundamental}). To do that, two metrics (or scores) can and should be used. 
    
    \subsubsection{Logarithimic Overfitting Ratio}
    \label{sec:lor}

    A note on notation is in order. When we write MAE, we intend the average of the single MAEs on each fold (if you are doing $k$-fold CV) or the single MAEs from each split in Monte-Carlo CV\footnote{Note that since each fold or split has the same number of elements the average of the fold averages is equal to the total average of the absolute errors.}.

    The Logarithimic Overfitting Ratio (LOR) is defined by (we will use the MAE as metric as an example, but it can also be defined with the MSE for example) the following.
    \begin{equation}
        \text{LOR} = \log \frac{\text{MAE}_\text{train}}{\text{MAE}_\text{test}}
    \end{equation}
    Its value can be interpreted as follows.
    \begin{itemize}
        \item $\mathbf{\textbf{LOR}=0}$: Ideal model. No overfitting, as the MAE on the train and on the test is the same.
        \item $\mathbf{\textbf{LOR}<0}$: Overfitting is present. The higher the value of LOR (in absolute value), the larger the overfitting.
        \item $\mathbf{\textbf{LOR}>0}$: There is underfitting. The higher the value of LOR (in absolute value), the larger the underfitting. 
    \end{itemize}
    \begin{defbox}{Logarithimic Overfitting Ratio (LOR)}
        The Logarithimic Overfitting Ratio (LOR) is defined by (we will use the MAE as metric as an exmaple, but it can be also defined with the MSE for example) the following.
    \begin{equation}
        \text{LOR} = \log \frac{\text{MAE}_\text{train}}{\text{MAE}_\text{test}}
    \end{equation}
    \end{defbox}
    You should choose the model with the LOR value closest to zero.
    The LOR is a nice metric, but it does not take into account the spread of the values over folds. Overfitting rears its ugly head in subtle ways. One is that although the metric average could be very similar on the training and test datasets, their standard deviation ($\sigma_\text{test}$ and $\sigma_\text{train}$) around the averages may be quite different. Typically, in overfitting regimes, we observe $\sigma_\text{test}\gg\sigma_\text{train}$. It is important to choose a model instance that not only shows similar metric averages but also a similar standard deviation. To facilitate the choice, you should use, in addition to the LOR, the \textbf{Composite Overfitting Ratio} (COS) defined by
    \begin{equation}
        \text{COS} = \alpha \frac{\text{MAE}_\text{train}}{\text{MAE}_\text{test}}+ \beta \frac{\sigma_\text{train}}{\sigma_\text{test}}
    \end{equation}
    The value of $\alpha$ and $\beta$ can, in principle, be freely chosen, but I suggest you use $\alpha=\beta=1/2$. The COS values can be interpreted this way.
    \begin{itemize}
        \item $\mathbf{\textbf{COS}=1}$: Optimal model. Train and test errors match, and the variance across folds (or splits in MC CV) is stable. Thus, the model shows good generalisation properties.
        \item $\mathbf{\textbf{COS}>1}$: Overfitting is present and there is some instability across folds (or splits in MC CV). The higher the value, the worse the situation.
        \item $\mathbf{\textbf{COS}<1}$: There is underfitting. The higher the value, the worse the situation.
    \end{itemize}
    \begin{defbox}{Composite Overfitting Ratio (COS)}
        the Composite Overfitting Ratio (COS) is defined by
    \begin{equation}
        \text{COS} = \alpha \frac{\text{MAE}_\text{train}}{\text{MAE}_\text{test}}+ \beta \frac{\sigma_\text{train}}{\sigma_\text{test}}
    \end{equation}
    with $\alpha,\beta >0$.
    \end{defbox}

\begin{figure}[hbt]
        \centering
        \includegraphics[width=1\linewidth]{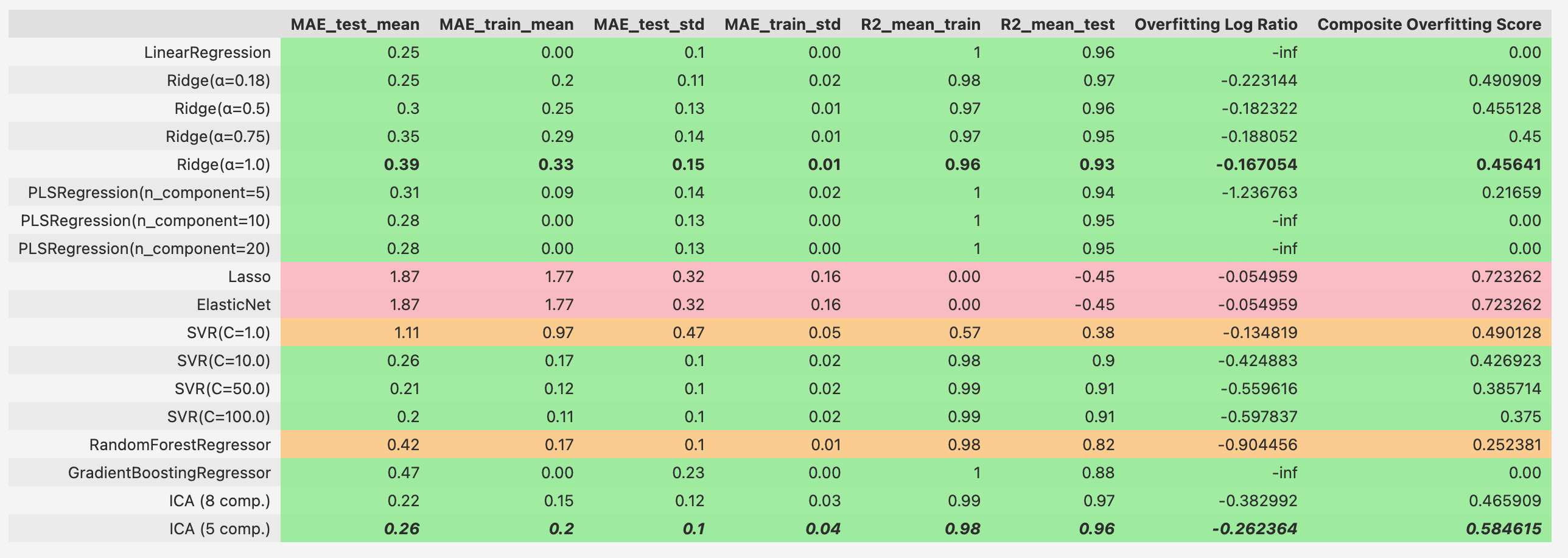}
        \caption{An example on how a table summarising results of experiments might look like. Note that this table reports only metrics and no other information, since in the project we generated one table for each experiment. But it should give you an idea about how it could look like. The red lines are those for which $R^2<0$, yellow those for which $R^2<0.85$ and green ones for which $R^2>0.85$. Model instnaces that have $R^2>0.85$ are the one that should be considered. In simple bold face we have highlighted the model instance for the LOR closest to zero, while in bold face and italics the one with the COS closest to 1. }
        \label{fig:example_table}
    \end{figure}

    In Figure \ref{fig:example_table} you can see how such a table might look like for a real project. Note that this table reports only metrics and no other information, since in the project we generated one table for each experiment. But it will give you an idea on how to report results systematically. The red lines are those for which $R^2<0$ (thus unusable), yellow those for which $R^2<0.85$ (therefore bad) and green ones for which $R^2>0.85$ (possibly useful model instances). The model instances that have $R^2>0.85$ are the ones that should be considered. In simple bold face, we have highlighted the model instance with the LOR closest to zero, while in bold face and italics the one with the COS closer to 1.

    \subsection{Extending to Deep Learning}

    If you are testing deep learning, you should optimise the network's architecture and then only report the one that gives you the best results. Unless your focus is on discussing hyperparameters.

    \begin{tipsbox}{Deep Learning}
    \begin{enumerate}[label=(\roman*)]
        \item Start with minimal architectures (1–2 hidden layers for FFNN, shallow CNN for images).
        \item Use early stopping, dropout, and batch normalization to control overfitting.
        \item Maintain consistent data splits and preprocessing between ML and DL models.
        \item Compare neural network results fairly against classical baselines.
    \end{enumerate}
    \end{tipsbox}

    \subsection{Selecting Meaningful Results}

    You should always report all your findings, but in deep learning sometime you find yourself in a situation where a model does not learn anything (something you can easily check by looking at the plot of the loss function vs. epochs). In this case the model instance result should not be reported. This manifest itslef often in a model instance predicting blindly always one of the classes (for example in a classification task). You can also see that by checking the confusion matrices in case you are working on a classification problem. In a regression problem, this manifest in a model instance that predict always almost the same value.
    
    \begin{tipsbox}{Selecting meaningful results}
    \begin{enumerate}[label=(\roman*)]
        \item Exclude models that clearly overfit or fail to learn (e.g., constant predictions).
        \item Include training/test metrics and standard deviations in all reports.
        %\item Prefer interpretable baselines for presentations and publications.
    \end{enumerate}
    \end{tipsbox}

    \section{Result Table and Reporting}
    
    \begin{tipsbox}{Results}
    \begin{enumerate}[label=(\roman*)]
        \item Use a uniform metric across models (e.g., MAE or accuracy) for fair comparison.
        \item Highlight the best model(s) in bold or with visual cues.
        \item Include preprocessing details explicitly: they often explain differences in performance.
        \item Keep the table concise, but add references to detailed results in an appendix if needed.
    \end{enumerate}
    \end{tipsbox}

% \begin{table}[h!]
%     \centering
%     \setlength{\tabcolsep}{10pt} % Adjust column spacing
%     \renewcommand{\arraystretch}{1.2} % Adjust row height
%     \begin{tabular}{l l c c}
%         \toprule
%         \textbf{Model} & \textbf{Pre-processing Details} & \textbf{Training (MAE $\pm$ SD)} & \textbf{Test (MAE $\pm$ SD)} \\
%         \midrule
%         Linear Regression & Raw data & $3.0 \pm 0.3$ & $3.2 \pm 0.5$ \\
%         SVM & Normalization (max = 1) & $2.2 \pm 0.1$ & $5.6 \pm 0.5$ \\
%         \bottomrule
%     \end{tabular}
%     \vspace{0.1cm}\caption{Example of model performance. The metric reported is the Mean Absolute Error (MAE) but could something else for you, for example the MSE, or the accuracy, or something completely different.}
%     \label{tab:example}
% \end{table}

\section{Aknowledgments}

We would like to thank Michael W\"uest for the helpful review and feedback.
\bibliographystyle{unsrt}  
\bibliography{references}

\end{document}